\documentclass[10pt,twocolumn,letterpaper]{article}

\usepackage[pagenumbers]{cvpr} 

\definecolor{cvprblue}{rgb}{0.21,0.49,0.74}
\usepackage[pagebackref,breaklinks,colorlinks,allcolors=cvprblue]{hyperref}
\usepackage{hyperref, url, algorithm, algpseudocode, multirow, adjustbox, tabularx, booktabs, amsmath, amssymb, pifont, mathtools}

\usepackage{CJKutf8}    
\usepackage{colortbl}
\usepackage[table]{xcolor}
\usepackage[misc]{ifsym}

\definecolor{cvprblue}{rgb}{0.21,0.49,0.74}
\usepackage{float}
\usepackage[pagebackref,breaklinks,colorlinks,allcolors=cvprblue]{hyperref}

\def\confName{CVPR}
\def\confYear{2026}

\newcommand{\model}{WAM-Diff}
\title{\model: A Masked Diffusion VLA Framework with MoE and Online Reinforcement Learning for Autonomous Driving}

\newcommand{\authorskipshort}{\hspace{3mm}}
\author{
  \begin{tabular}{c}
  Mingwang~Xu$^{1*}$ \authorskipshort
  Jiahao~Cui$^{1*}$ \authorskipshort
  Feipeng~Cai$^{2*}$ \authorskipshort
  Hanlin~Shang$^{1*}$ \authorskipshort
  Zhihao~Zhu$^{1}$ \authorskipshort
  Shan~Luan$^{1}$ 
  \\ [1mm]
  Yifang~Xu$^{1}$ \authorskipshort
  Neng~Zhang$^2$ \authorskipshort
  Yaoyi~Li$^2$  \authorskipshort
  Jia~Cai$^2$  \authorskipshort
  Siyu~Zhu$^{1}\textsuperscript{\Letter}$
  \end{tabular}
  \\ [4mm]
  $^1$Fudan University \hspace{8mm}
  $^2$Yinwang Intelligent Technology Co., Ltd
}

\begin{document}
\maketitle

\begingroup
\renewcommand\thefootnote{}\footnote{
    $^*$: Equal contribution.  \hspace{18mm}   \Letter: Corresponding authors.
} 
\endgroup

\vspace{-7mm}

\begin{abstract}
End-to-end autonomous driving systems based on vision-language-action (VLA) models integrate multimodal sensor inputs and language instructions to generate planning and control signals. 
While autoregressive large language models and continuous diffusion policies are prevalent, the potential of discrete masked diffusion for trajectory generation remains largely unexplored. 
This paper presents \model, 
a VLA framework that employs masked diffusion to iteratively refine a discrete sequence representing future ego-trajectories. 
Our approach features three key innovations:
a systematic adaptation of masked diffusion for autonomous driving that supports flexible, 
non-causal decoding orders; scalable model capacity via a sparse MoE architecture trained jointly on motion prediction and driving-oriented visual question answering (VQA); 
and online reinforcement learning using Group Sequence Policy Optimization (GSPO) to optimize sequence-level driving rewards. 
Remarkably, our model achieves 91.0 PDMS on NAVSIM-v1 and 89.7 EPDMS on NAVSIM-v2, 
demonstrating the effectiveness of masked diffusion for autonomous driving. 
The approach provides a promising alternative to autoregressive and diffusion-based policies, supporting scenario-aware decoding strategies for trajectory generation. The code for this paper will be released publicly at: \url{https://github.com/fudan-generative-vision/\model}.
\end{abstract}

\begin{figure}[!t]
    \centering 
    \includegraphics[width=1.0\columnwidth]{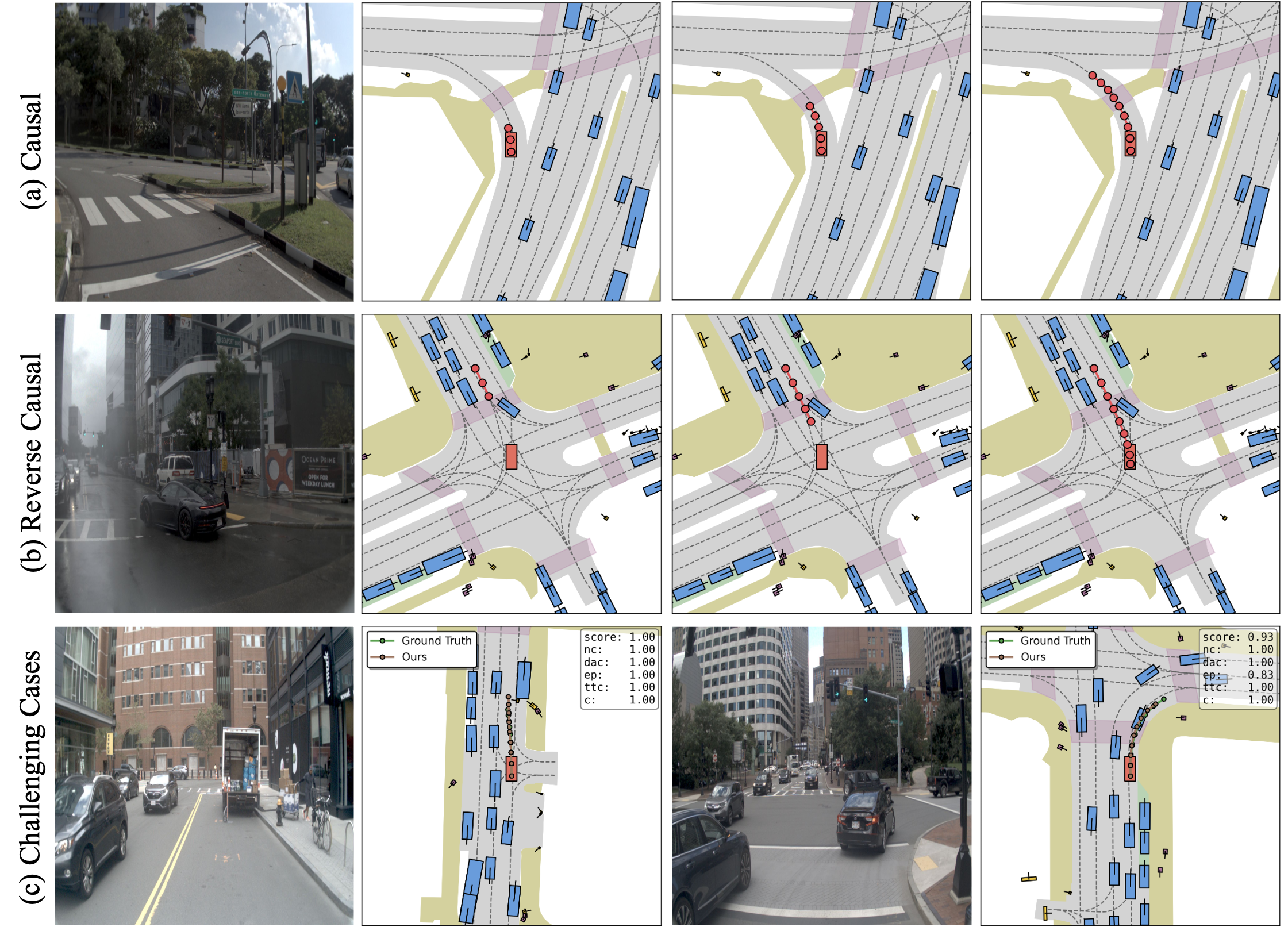} 
    \vspace{-3.5mm}
    \caption{The proposed \model\space framework supports flexible decoding orders for motion planning, 
    illustrated for (a) causal, (b) reverse-causal, and random schedules, adapting to diverse driving scenarios. 
    (c) By integrating a Mixture-of-Experts architecture with GSPO reinforcement learning, the model achieves superior performance in challenging driving situations.}
    \label{fig:teaser}
    \vspace{-6mm}
\end{figure}



    
\begin{figure*}[!ht]
  \centering
  \includegraphics[width=\textwidth]{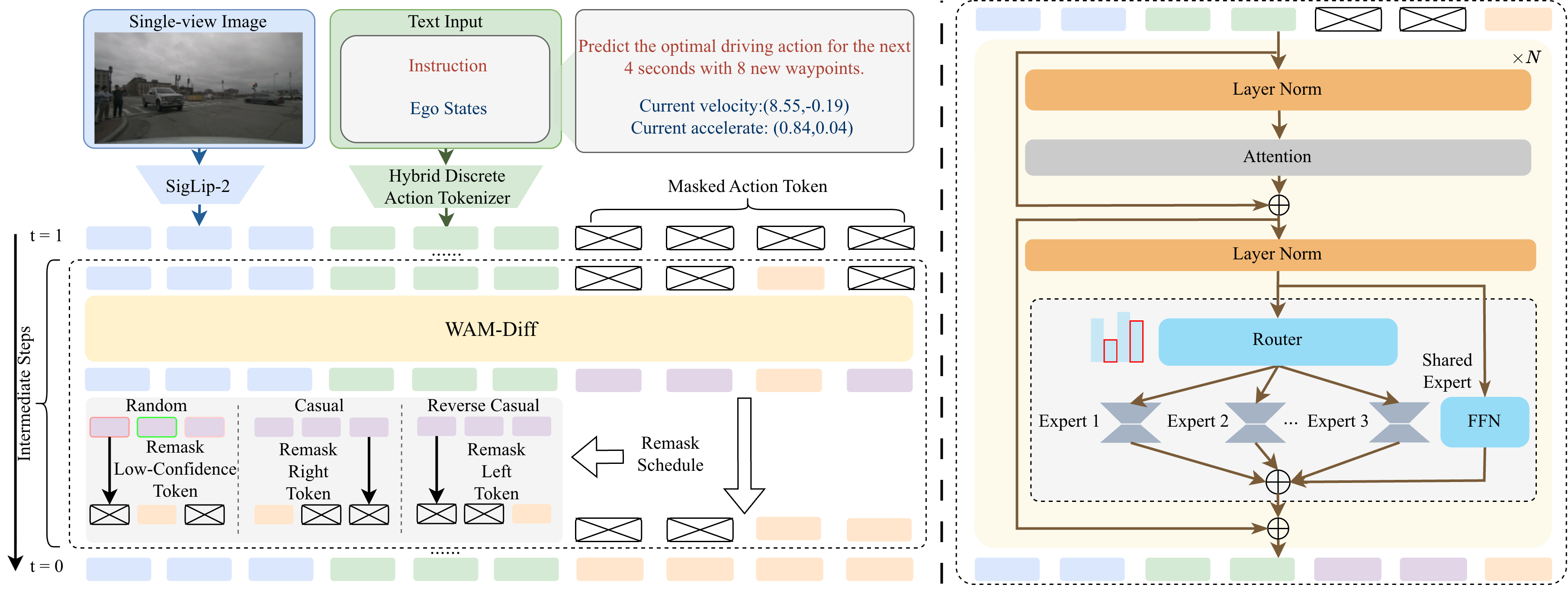} 
  \vspace{-4 mm}
  \caption{Overview of \model, the proposed VLA framework for end-to-end autonomous driving. 
  The architecture integrates a MoE-enhanced backbone with a discrete mask diffusion decoder. 
  Multimodal inputs--including ego camera images, ego-states, navigation, instructions--are encoded and fused. 
  The masked diffusion process then iteratively generates the future trajectory (represented as a sequence of waypoints) from a fully masked initial state, 
  guided by a remasking scheduler. 
  The overall pipeline supports joint training with supervised multi-task learning and online GSPO reinforcement learning for improved driving performance.}
  \label{fig:pipeline}
  \vspace{-4mm}
\end{figure*}

\vspace{-1mm}
\section{Introduction}
\label{sec:intro}

End-to-end autonomous driving systems~\cite{kim2024openvla, hwang2024emma, mao2023gpt, xing2025openemmaopensourcemultimodalmodel,mao2023language,wang2024omnidrive, bai20243d, shao2024lmdrive, zhang2024wisead, zhao2025sce2drivex} grounded in VLA paradigms~\cite{zhou2025autovla, zhou2025opendrivevla, arai2025covla, kim2024openvla, wen2025lladavlavisionlanguagediffusion} aim to integrate natural-language instructions with rich multi-sensor perceptual data.
The objective is to develop unified frameworks capable of generating reliable control and planning signals directly from multimodal inputs.
By jointly modeling visual perception, linguistic understanding, and action generation, 
these end-to-end models signify a substantial shift from traditional modular pipelines towards more unified architectures suited for complex, dynamic, and open-ended traffic environments.
Current VLAs for autonomous driving primarily follow two architectural paradigms. 
The first encompasses autoregressive LLM-based approaches~\cite{hwang2024emma, xing2025openemmaopensourcemultimodalmodel, li2025recogdrive, zhou2025autovla}, which generate action sequences token-by-token, often leveraging extensive multimodal pretraining for strong generalization~\cite{kim2024openvla,zhou2025autovla,li2025recogdrive}. 
The second paradigm consists of diffusion policy models~\cite{nie2025large,li2025discrete}, 
which iteratively refine action predictions through a noise-to-target denoising process, offering an alternative for capturing complex multi-modal distributions.

Recently, discrete masked diffusion~\cite{nie2025large, wen2025lladavlavisionlanguagediffusion} has emerged as a promising generative architecture for sequential data, including language and multimodal tasks. 
This approach formulates sequence generation as an iterative infilling process: beginning with a fully masked sequence, the model progressively predicts all masked tokens in parallel at each step, while selectively re-masking low-confidence predictions.
This allows the model to leverage bidirectional context throughout the decoding process, 
overcoming the inherent left-to-right generation constraint of autoregressive models. 
Such a paradigm is particularly suitable for trajectory generation in autonomous driving, as it naturally supports flexible decoding orders that can incorporate scenario-specific priors. For example, a causal order is well suited to near-term maneuvering such as turning, a reverse-causal order benefits car-following or oncoming interactions that require long-range anticipation, and a random order provides a balanced default.
Nevertheless, its application to autonomous driving remains underexplored; this work addresses that gap.

This paper presents a systematic exploration of masked diffusion for autonomous driving VLAs, organized around three principal contributions.
First, we conduct an in-depth analysis of the masked diffusion architecture adapted to the driving context. 
This includes the design of a hybrid discrete action tokenization scheme that interleaves numerical trajectory waypoints with semantic language tokens, 
improving the precision of future action prediction compared to purely text-based representations. Capitalizing on the inherent flexibility of masked diffusion, 
we investigate different decoding schedules for generating future action sequences and analyze their respective suitability for various driving scenarios.
Second, we scale capacity with a sparse MoE via LoRA MoE~\cite{dou2024loramoe1,zhu2025llada}, 
yielding a 64-expert masked diffusion backbone that supports scalable motion planning.
We demonstrate that joint training on both motion prediction and driving-oriented VQA tasks within this MoE framework yields superior performance compared to motion-only training, enhancing the model's motion planning capabilities.
Third, we incorporate the online reinforcement learning GSPO, tailored for the MoE framework. 
This approach optimizes the policy against multi-dimensional reward signals from simulation (e.g., non-collision, comfort, ego-progress), 
leading to significant improvements in overall driving performance as measured by standard benchmarks.

Experiments on the NAVSIM-v1 and v2 benchmarks~\cite{dauner2024navsim} demonstrate the effectiveness of the proposed approach. 
To the best of our knowledge, this is the first VLA for autonomous driving that integrates discrete masked diffusion with a sparse MoE and online GSPO reinforcement learning. 
Specifically, the model achieves 91.0 PDMS on NAVSIM-v1 and 89.7 EPDMS on NAVSIM-v2, reaching leading autoregressive baselines, underscoring the promise of masked diffusion decoders for VLAs. 
Beyond accuracy, masked diffusion circumvents the inherent left-to-right constraint of autoregressive generation, enabling random, casual, reverse-casual and scenario-aware decoding schedules.
Through experiments, we demonstrate the flexibility of our method in generating trajectories with different decoding orders tailored to specific scenario priors.

\begin{figure}[!t]
  \centering
  \includegraphics[width=\linewidth]{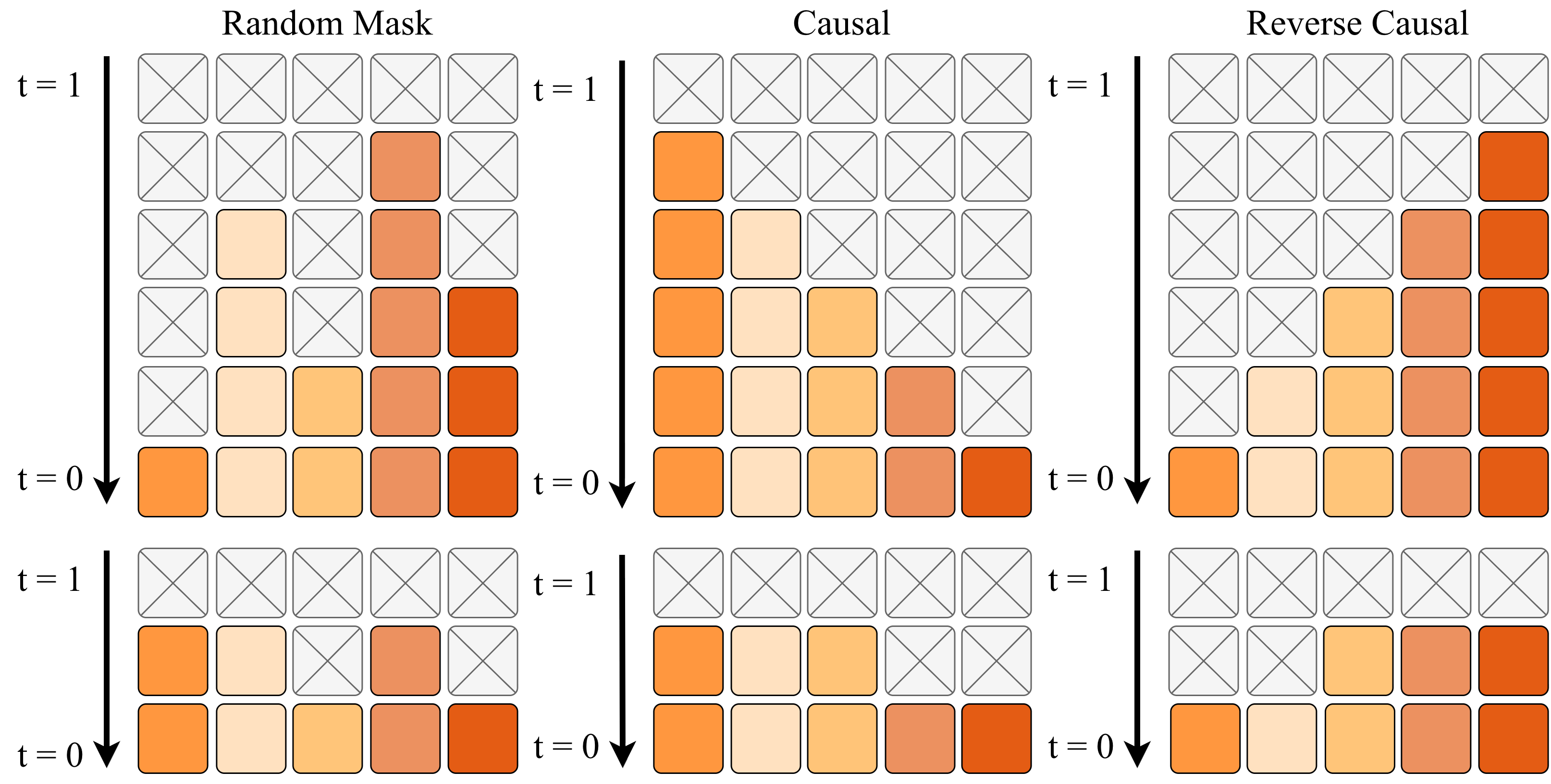} 
  \vspace{-4mm}
  \caption{Decoding schedules for masked diffusion. 
  Top figures: remasking policies that regulate trajectory‑token update order (random, causal, reverse‑causal).
  Button figures: decoding efficiency via the mask‑rate schedule.}
  \label{fig:decode_scheduler}
  \vspace{-3mm}
\end{figure}
\section{Related Work}
\label{sec:relwork}

\noindent\textbf{End-to-End Autonomous Driving.}
Recent advances emphasize jointly optimizing perception and planning. 
UniAD~\cite{hu2023planning} integrates multiple perception tasks to improve planning quality; 
VAD~\cite{jiang2023vad} introduces compact vectorized scene representations, 
and VADv2~\cite{chen2024vadv2} extends this to multi-modal planning via trajectory scoring and anchor-based sampling. 
HydraMDP~\cite{li2024hydra} further stabilizes planning with rule-based supervision. 
ParaDrive~\cite{weng2024drive} analyzes core design choices for end-to-end systems, 
while generative approaches such as GenAD~\cite{zheng2024genad} and DiffusionDrive~\cite{liao2025diffusiondrive} leverage generative and diffusion modeling to capture multi-modal, 
temporally coherent trajectories. 
In parallel, vision–language models for driving have evolved from interpretive systems that describe scenes without direct control~\cite{xu2024drivegpt4,tian2024drivevlm}, 
through modular language-to-action pipelines with non-differentiable interfaces~\cite{zhou2025opendrivevla,yuan2024rag,arai2025covla,zhang2025safeauto}, 
to unified VLA architectures--such as DriveMoE~\cite{yang2025drivemoe}, ReCogDrive~\cite{li2025recogdrive}, AutoVLA~\cite{zhou2025autovla}--that map sensory inputs to trajectories within a single differentiable model. 
We follow this paradigm and investigate discrete masked diffusion as a bidirectionally conditioned, parallel decoder for trajectory generation in end-to-end autonomous driving.

\noindent\textbf{Discrete Diffusion in Multimodal LLMs.}
Early studies such as D3PM~\cite{austin2021structured} and SEDD~\cite{SEDD-2023} established the foundation for diffusion over discrete variables. Recent approaches generally follow two denoising paradigms. 
The first, the masked diffusion process, formulates generation as iterative infilling: starting from a fully masked sequence, the model progressively predicts tokens in parallel and re-masks uncertain ones, enabling flexible decoding orders through an absorbing-state mechanism. 
LLADA~\cite{nie2025large} and DREAM~\cite{DREAM-2025} extended this approach to large-scale text generation, while LLADA-V~\cite{llada-V-2025} and MMADA~\cite{MMADA-2025} further expanded it to multimodal learning, demonstrating strong cross-modal reasoning and consistency.
In contrast, uniform diffusion models~\cite{austin2021structured,campbell2022continuous,gat2024discrete,FUDOKI-2025} employ structured transition kernels between discrete states, offering a probabilistic framework for sequence generation and showing strong performance in multimodal consistency tasks.
Although, discrete diffusion models have emerged as a promising alternative to autoregressive generation for sequential data,
preliminary attempts to apply discrete diffusion to autonomous driving~\cite{cui2025vilad,li2025discrete} exhibit performance gaps compared to state-of-the-art autoregressive methods.
In this paper we aim to resolve this gap by scaling masked diffusion with MoE and GSPO online reinforcement learning,
and leverage its advantages in parallel decoding and bidirectional context modeling~\cite{yu2025-dllm-survey}, 

\begin{figure}[!t]
  \centering
  \includegraphics[width=\linewidth]{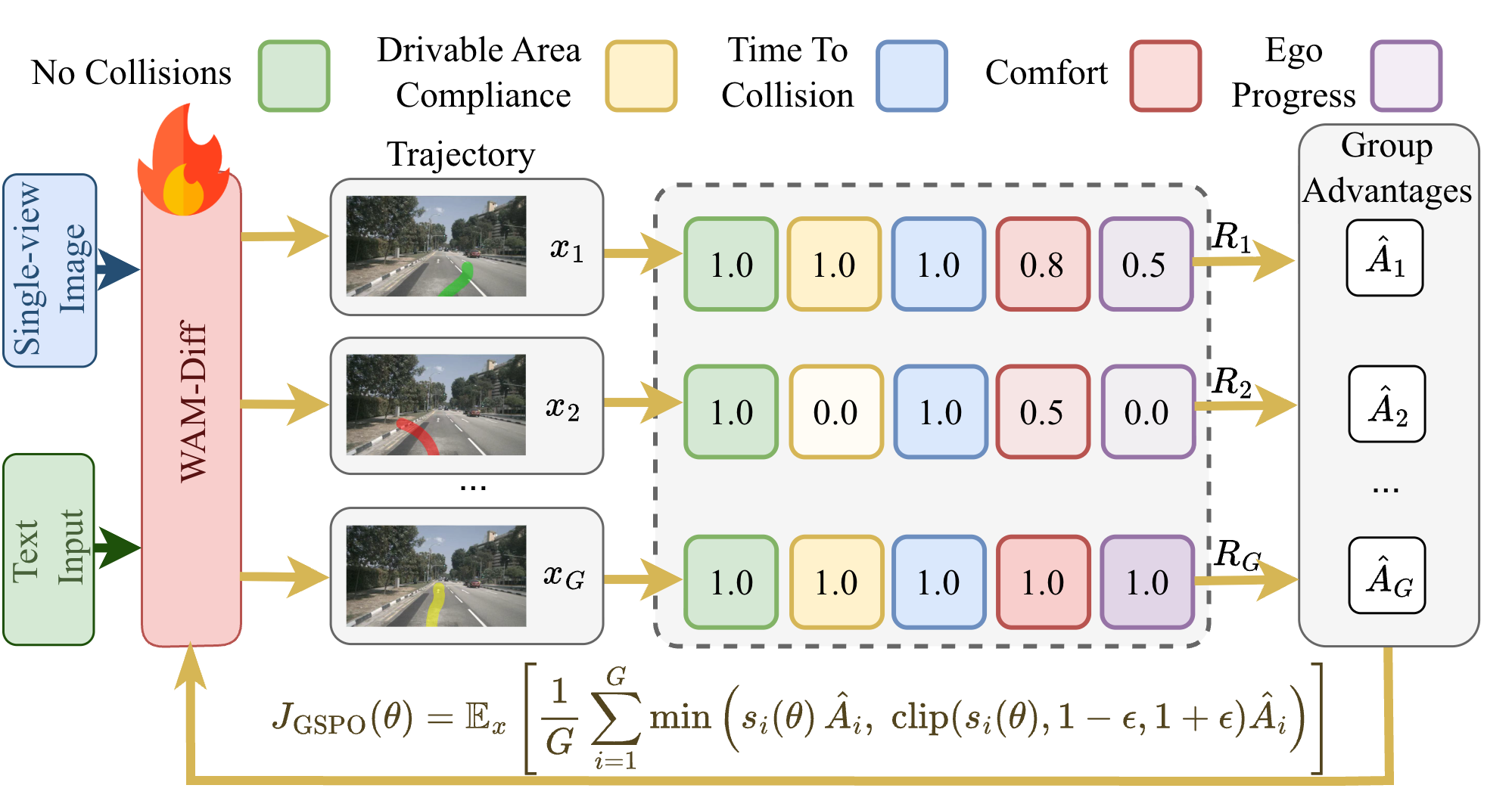} 
  \vspace{-5mm}
  \caption{Illustration of GSPO integrating multi-factor safety rewards—no collisions, drivable-area compliance, time-to-collision, comfort, and ego progress—into masked diffusion trajectory optimization for end-to-end autonomous driving.}
  \label{fig:pipeline}
  \vspace{-3.5mm}
\end{figure}

\noindent\textbf{Mixture of Experts in VLAs.}
The MoE architecture is a principal mechanism for parameter‑efficient scaling and task specialization in large language and vision–language models via dynamic expert routing~\cite{DeepseekMoe-2024, GeminiMoe-2024, Qwen3-2025}. 
Recent VLA systems leverage MoE to disentangle visuomotor control from vision–language reasoning: 
ChatVLA~\cite{CHATVLA-2025} shares attention while separating feed‑forward pathways, 
and ChatVLA2~\cite{zhouchatvla} introduces dynamic MoE layers without explicit FFNs. 
MoRE~\cite{zhao2025more} transforms dense networks into MoE by inserting LoRA‑based experts, 
akin to LoRAMoE~\cite{dou2024loramoe1}, 
demonstrating scalable specialization across tasks; 
ForVLA~\cite{yu2025forcevla} adopts a standard MoE design for manipulation. 
In autonomous driving, DriveMoE~\cite{yang2025drivemoe} routes between scene‑ and skill‑specialized experts for view selection and maneuver generation, 
while ARTEMIS~\cite{feng2025artemisautoregressiveendtoendtrajectory} employs an autoregressive MoE planner for scene‑conditioned waypoint prediction. 
Building on these insights, 
we integrate a sparse LoRA‑based MoE into a masked diffusion backbone, 
enabling scalable adaptability across diverse and complex driving scenarios.

\section{Method}
\label{sec:method}
We cast end-to-end autonomous driving as conditional masked diffusion over a unified discrete sequence that encodes future ego-trajectory under multimodal perceptual context and language guidance (Section~\ref{subsec:problem}). 
Building on this formulation, we instantiate masked diffusion for VLA, contrast it with autoregressive left-to-right decoding and diffusion policies, 
and exploit confidence- and prior-aware remasking schedules (causal, reverse-causal, and random) to control decoding order and efficiency for trajectory generation (Section~\ref{subsec:maskdiff}). 
To scale capacity and semantics, 
we integrate a sparse LoRA-based MoE into the diffusion backbone and jointly train on motion prediction and driving-oriented VQA, yielding stronger motion planning. (Section~\ref{subsec:moe}). 
We further incorporate online GSPO to optimize sequence-level rewards for safety, ego progress, and comfort (Section~\ref{subsec:moe}). 
Finally, we present the overall architecture together with training and inference procedures (Section~\ref{subsec:architecture}).

\subsection{Problem Formulation}
\label{subsec:problem}
We cast end-to-end autonomous driving as conditional sequence modeling over a unified discrete representation of future actions. 
At each decision step $t$, the agent observes a multimodal context $c_t = \{I_t, s_t, u_t\}$, 
where $I_t$ denotes single-view camera images, $s_t$ the ego-vehicle state (e.g., velocity, acceleration, navigation), 
and $u_t$ a natural-language instruction.
The prediction target is a future action sequence 
$x_0 = (x_0^1,\dots,x_0^L)$ 
of length $L$, 
which encodes a planned ego-trajectory as a sequence of tokens drawn from a shared vocabulary spanning both numerical (metric waypoints) and semantic (control/rationale) symbols. 
Our objective is to learn the conditional distribution $p_\theta(x_0 \mid c_t)$. 
We instantiate $p_\theta$ with a discrete masked diffusion decoder that starts from a fully masked sequence and iteratively infills tokens under the joint conditioning of visual, state, and linguistic inputs.

\subsection{Masked Diffusion for VLA}
\label{subsec:maskdiff}

\noindent\textbf{Hybrid Discrete Action Tokenization.}
We construct a unified vocabulary capable of representing both numerical values and textual tokens. 
Continuous variables—such as trajectory waypoints—are uniformly quantized over the interval $[-100, 100]$ with a resolution of $0.01$, 
resulting in $20{,}001$ distinct numerical tokens. 
Each 2D waypoint is represented as an ordered pair $\langle x, y \rangle$ of scalar tokens; 
during decoding, the bin center of each quantized token is used, introducing a maximum absolute error of $0.005$ per coordinate. 
Semantic control commands and driving rationales (e.g., lane-keep, yield, turn-left) are represented using their corresponding textual tokens. 
The $20{,}001$ numerical tokens are merged into the existing text vocabulary, 
and their embedding projections are optimized end-to-end during masked diffusion training. 
This hybrid tokenization supports seamless interleaving of metric and linguistic information, 
enabling bidirectional conditioning while preserving both numerical precision and semantic interpretability.

\noindent\textbf{Masked Diffusion for Trajectory Generation.}
Let $M$ denotes the special mask token and $r\in[0,1]$ the mask rate. 
The forward corruption independently replaces each position with $M$ with probability $r$:
\begin{equation}
\label{eq:forward-app}
q_r(x_r | x_0)
= \prod_{i=1}^{L}\Big[(1-r)\mathbf{1}\{x_r^i=x_0^i\} + r\mathbf{1}\{x_r^i=M\}\Big].
\end{equation}
At inference, 
the reverse model $p_\theta(\cdot | x_r, c_t)$ jointly predicts all masked tokens conditioned on the visible tokens and context. 
A confidence-based remasking policy then re-masks low-confidence predictions to form a new corrupted sequence, 
and this infill–remask procedure is iterated until all tokens are resolved. 
This yields globally consistent, parallel decoding in a small number of steps.

For training, we minimize a masked cross-entropy with a continuous mask rate $r\sim\mathcal{U}(0,1)$, 
normalized by the expected number of masked tokens:
\begin{equation}
\label{eq:loss-compute}
\mathcal{L}(\theta)
= -\mathbb{E}_{x_0, r, x_r}\Bigg[
\frac{1}{r}\sum_{i=1}^{L}\mathbf{1}\{x_r^i=M\}
\log p_\theta\big(x_0^i | x_r, c_t\big)
\Bigg].
\end{equation}
This objective upper-bounds the negative log-likelihood of 
$p_\theta(x_0 | c_t)$
is Fisher-consistent in the large-data limit, 
and enables bidirectional dependencies across visual and textual tokens.

The masked diffusion formulation offers two advantages for VLAs: 
1) Parallel, global prediction at each step enables efficient generation, potentially converging in few steps. 
2) Controllable decoding via remasking schedules can inject driving priors into the token update order, and therefore surpass the causal constraints of autoregressive decoders.

\noindent\textbf{Decoding Schedules for Action Sequences.}
At inference, we employ a decreasing mask schedule defined by 
$1 = r_0 > r_1 > \cdots > r_T = 0$, 
beginning from a fully masked sequence $x_{r_0} = M^L$. 
For each step $j = 0, \dots, T-1$, 
the model performs two operations: 
it first infills all currently masked tokens by sampling from the conditional distribution 
$p_\theta(\cdot | x_{r_j}, c_t)$; 
subsequently, a remasking policy is applied to selectively re-mask a subset of tokens, 
yielding the input for the next step, $x_{r_{j+1}}$. 
The design of the remasking policy allows optimization of the masked diffusion process along two principal dimensions: the decoding order and the decoding efficiency.

As shown in Figure~\ref{fig:decode_scheduler}, the decoding order can be aligned with driving priors: 
a confidence-driven \textbf{Random} schedule re-masks low-confidence tokens irrespective of time; 
a \textbf{Causal} schedule monotonically unmasks tokens in temporal order to promote kinematic coherence; 
and a \textbf{Reverse-causal} schedule resolves distant future tokens before near-term ones to first stabilize long-horizon intent and subsequently refine immediate actions.

The decoding efficiency is simultaneously regulated by the remasking rate, 
which determines the fraction of tokens fixed per iteration, 
ranging from fine-grained single-token updates to full-sequence infilling in one or a few steps. 
Appropriate choices of order and efficiency yield faster convergence and improved trajectory accuracy and consistency across scenarios.

\subsection{Scaling Masked Diffusion with LoRA MoE}
\label{subsec:moe}

Purely trajectory-supervised training equips the model to mimic motion patterns, 
but it is insufficient for safe and efficient driving that requires semantic scene understanding, traffic-rule compliance, and interactive reasoning. 
To address these limitations, we enhance our framework through two complementary approaches: 
1) integrating a MoE architecture with multi-task learning combining trajectory prediction and driving-oriented VQA;
2) employing online reinforcement learning, namely GSPO, to further optimize safety, progress, and comfort metrics.

\noindent\textbf{LoRA MoE Enhanced Masked Diffusion VLA.}
We incorporate a sparse LoRA MoE architecture into the feed-forward networks of our masked diffusion backbone to enable specialized expert capacity for different driving scenarios. 
The LoRA MoE formulation maintains parameter efficiency while allowing flexible scaling to accommodate complex driving situations. 
Formally, for an input representation $z$, the output of a LoRA MoE layer with $N$ experts is given by:
\begin{equation}
o = W_0 z + \sum_{i=1}^{N} g_i(z) E_i(z),
\end{equation}
where $W_0$ denotes the pre-trained feed forward network (FFN) projection matrix, 
$E_i(z) = B_i A_i z$ represents the low-rank adaptation of the 
$i$-th expert (with $A_i \in \mathbb{R}^{r \times d_{\text{in}}}$, 
$B_i \in \mathbb{R}^{d_{\text{out}} \times r}$, 
and rank $r \ll \min(d_{\text{in}}, d_{\text{out}}))$, 
and $g_i(z)$ is the routing weight produced by a softmax gate $g(z) = \text{Softmax}(W_g z)$.

This architecture enables dynamic routing of inputs to specialized experts based on scenario characteristics. 
We train the MoE-enhanced model on a hybrid objective combining trajectory prediction and driving-oriented VQA tasks. 
This allows the model to learn motion prediction capabilities not only through data-driven trajectory imitation but also through visual instruction training. 
This further enhances its capabilities from low-level scene perception and motion prediction to advanced driving planning and decision-making, thereby improving the model's comprehensive ability to predict motion across multiple dimensions.

\noindent\textbf{GSPO for MoE Masked Diffusion.}
After multi-task supervised MoE pretraining, 
we further online reinforce the masked diffusion policy with GSPO specifically for safety, ego progress, and comfort, 
which optimizes rewards at the level of whole action sequences and is therefore well-suited to sparse-expert routing. 

For each context $c_t$ of a single driving process,
we sample a group of $G$ candidate sequences 
$\{x_i\}_{i=1}^G \sim \pi_{\theta_{\text{old}}}(\cdot|c_t)$. 
Each sequence is evaluated by the NAVSIM simulator under the PDMS metric, 
yielding rewards $R_i=r(c_t,x_i)$. 
GSPO converts these rewards into a group-normalized advantage,
that is:
$\hat A_i=\frac{R_i-\mathrm{mean}(\{R_j\}_{j=1}^G)}{\mathrm{std}(\{R_j\}_{j=1}^G)}$,
which reduces scale sensitivity and variance across prompts.

Policy improvement is driven by a sequence-likelihood importance ratio that is length-normalized, for efficiency, we use one-step unmasking to estimate per-token log-probability $log \pi_\theta (x_{i,k}|c_t)$ following \cite{zhao2025d1}, and the importance ratio is calculated as:
{\small
\begin{equation}
s_i(\theta)=\left(\frac{\pi_{\theta}(x_i|c_t)}{\pi_{\theta_{\text{old}}}(x_i|c_t)}\right)^{\frac{1}{|x_i|}}
=\exp\left(\frac{1}{|x_i|}\sum_{k=1}^{|x_i|}\log\frac{\pi_{\theta}(x_{i,k}|c_t)}{\pi_{\theta_{\text{old}}}(x_{i,k}|c_t)}\right),
\end{equation}}
where we compute sequence likelihood under a fixed factorization of the hybrid token sequence for comparability across policies. 
The GSPO objective then mirrors PPO at the sequence level:   
{\small
\begin{equation}
\label{eq:gspo}
J_{\text{GSPO}}(\theta)=
\mathbb{E}_{x}\left[
\frac{1}{G}\sum_{i=1}^{G}
\min\Big(s_i(\theta)\hat A_i, \mathrm{clip}\big(s_i(\theta),1-\epsilon,1+\epsilon\big)\hat A_i\Big)
\right].
\end{equation}}
We maximize $J_{\text{GSPO}}$ (equivalently minimize $-J_{\text{GSPO}}$) with $\theta_{\text{old}}$ held fixed during an update epoch, 
and periodically set $\theta_{\text{old}}\leftarrow\theta$.

Optimizing entire sequences avoids token-wise credit assignment and the associated instability from changing expert routes,
which is a severe problem for GRPO.
Consequently, GSPO provides stable credit to the MoE policy for trajectories that score well under NAVSIM simulation, 
while the clipping term constrains policy updates measured in sequence likelihood space, 
yielding robust improvements in both safety and closed-loop driving performance.

\subsection{Network Architecture}
\label{subsec:architecture}
\noindent\textbf{Overall Structure.}
The proposed architecture integrates four core components for multimodal reasoning and trajectory generation. 
The image encoder processes raw camera inputs by partitioning the 1920×1080 image into 15 non-overlapping 384×384 patches while simultaneously resizing the full image to the same reduced resolution. 
These 16 patches are encoded using SigLIP-2, yielding 2,185 visual tokens that are projected to 4,096 dimensions via an MLP to align with the text feature space. 
The text encoder extends the original vocabulary of 126,349 tokens by 20,001 new tokens for quantized waypoint representation, 
resulting in a total vocabulary size of 146,350. Training instances are structured using fixed question-answer templates that incorporate navigation commands, ego-state information, and waypoint predictions.




\begin{figure*}[!t]
  \centering
  \includegraphics[width=\textwidth]{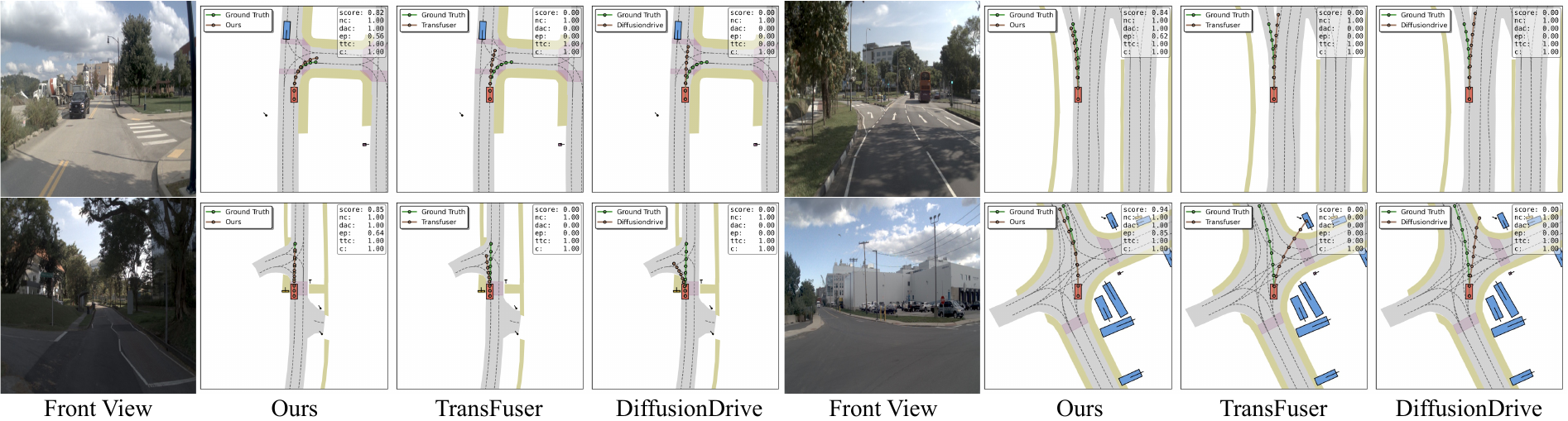} 
  \vspace{-6mm}
  \caption{Qualitative comparison with existed open-sourced methods on NAVSIM benchmark.}
  \vspace{-3mm}
  \label{fig:vis_compare}
\end{figure*}

\begin{table*}[!t]
\centering
\begin{minipage}[t]{0.57\textwidth}
    \centering
    \small
    \setlength{\tabcolsep}{4pt}
    \resizebox{\linewidth}{!}{
    \begin{tabular}{l|c|c|c|c|c|c}
    \hline
    \textbf{Method} & \textbf{NC}$\uparrow$ & \textbf{DAC}$\uparrow$ &
    \textbf{TTC}$\uparrow$ & \textbf{Comf.}$\uparrow$ &
    \textbf{EP}$\uparrow$ & \textbf{PDMS}$\uparrow$\\
    \hline
    UniAD~\cite{hu2023planning} & 97.8 & 91.9 & 92.9 & \textbf{100} & 78.8 & 83.4 \\
    PARA-Drive~\cite{weng2024drive}& 97.9 & 92.4 & 93.0 & 99.8 & 79.3 & 84.0 \\
    TransFuser~\cite{chitta2022transfuser} & 97.7 & 92.8 & 92.8 & \textbf{100} & 79.2 & 84.0\\
    DRAMA~\cite{yuan2024drama} & 98.0 & 93.1 & 94.8 & \textbf{100} & 80.1 & 85.5\\
    VADv2-$\mathcal{V}$8192~\cite{chen2024vadv2} & 97.2 & 89.1 & 91.6 & \textbf{100} & 76.0 & 80.9 \\
    Hydra-MDP-$\mathcal{V}$8192~\cite{li2024hydra} & 97.9 & 91.7 & 92.9 &\textbf{100} & 77.6 & 83.0\\
    DiffusionDrive~\cite{liao2025diffusiondrive} & 98.2 & 96.2 & 94.7 & \textbf{100} & 82.2 & 88.1\\
    ReCogDrive~\cite{li2025recogdrive} & 97.9 &97.3& 94.9& \textbf{100} &\textbf{87.3} &90.8\\
    DriveVLA-W0~\cite{li2025drivevla}&98.7& 99.1& 95.3& 99.3& 83.3& 90.2\\
    \hline
    \textbf{Ours} &\textbf{99.1}  &\textbf{98.3}  &\textbf{96.5}&99.9   &84.4&\textbf{91.0} \\
    \hline
    \end{tabular}
    }
    \vspace{-1mm}
    \caption{Comparison with state-of-the-art methods on the NAVSIM-v1.}
     \vspace{-4mm}
    \label{tab:navsim_v1}
\end{minipage}
\hfill
\begin{minipage}[t]{0.42\textwidth}
    \centering
    \small
    \vspace{-2.19cm}
    \setlength{\tabcolsep}{4pt}
    \resizebox{\linewidth}{!}{
    \begin{tabular}{l|c|c|c|c|c|c}
    \hline
    \textbf{Method} & \textbf{NC}$\uparrow$ & \textbf{DAC}$\uparrow$ & \textbf{TTC}$\uparrow$ & \textbf{Comf}.$\uparrow$ & \textbf{EP}$\uparrow$ & \textbf{PDMS}$\uparrow$\\
    \hline
    w/o&97.8&94.2&93.4&99.7&78.5&84.7 \\
    \hline
    expert 16&98.0 &94.0  &93.5  &99.7 &79.0  &85.0 \\
    expert 64&98.0&\textbf{95.5}&\textbf{94.2}&\textbf{99.4}&\textbf{80.7}&\textbf{86.6}   \\
    \hline
    Rank 8&\textbf{98.2}  &94.4  &94.0   &99.0&79.5   &85.5  \\
    Rank 32&98.0&\textbf{95.5}&\textbf{94.2}&\textbf{99.4}&\textbf{80.7}&\textbf{86.6}  \\
    \hline
    \end{tabular}
    }
    \vspace{-1.8 mm}
    \caption{Ablation study on MOE configurations.}
    \vspace{1mm}
    \label{tab:moe_ablation}

    \small
    \setlength{\tabcolsep}{4pt}
    \resizebox{\linewidth}{!}{
    \begin{tabular}{l|c|c|c|c|c|c}
    \hline
    \textbf{Method} & \textbf{NC}$\uparrow$ & \textbf{DAC}$\uparrow$ & \textbf{TTC}$\uparrow$ & \textbf{Comf}.$\uparrow$ & \textbf{EP}$\uparrow$ & \textbf{PDMS}$\uparrow$\\
    \hline
    w/o&98.0&95.5&94.2&99.4&80.7&86.6\\
    GSPO (G=2)&98.4  &97.1  &94.9   &99.8   &83.2&88.9 \\
    GSPO (G=3)&\textbf{99.1}  &\textbf{98.3}  &\textbf{96.5}&\textbf{99.9}   &\textbf{84.4}&\textbf{91.0} \\
    \hline
    \end{tabular}
    }
    \vspace{-1 mm}
    \caption{Ablation study on online-reinforcement leaning GSPO. ``G'' denotes group sizes.}
    \vspace{-5mm}
    \label{tab:gspo_ablation}
\end{minipage}
\end{table*}

The core masked diffusion model builds upon the Llada-V multimodal backbone. 
To enhance capacity efficiently, we integrate a sparse MoE into the FFNs, 
incorporating 64 LoRA experts with rank 32. 
The routing follows an expert-choice strategy, 
with the original FFN parameters serving as a shared expert. 
This configuration results in a total of 8.4B parameters, 
with the MoE components adding only 0.5B parameters and activating approximately 0.05B during inference, 
thereby minimizing computational overhead. 
The text decoder consists of the original model's text head adapted to the expanded vocabulary, 
enabling generation of hybrid sequences that interleave linguistic rationales with discrete trajectory waypoints.

\noindent\textbf{Training.}
We adopt a four-stage training paradigm.
1) MoE warm-up.
The diffusion backbone is frozen while LoRA experts are trained for 0.2 epochs on 668K nuPlan trajectory samples to initialize planning capabilities and prevent mode collapse.
2) Multi-task supervised pre-training.
All parameters are unfrozen and jointly trained for one epoch on 668K trajectories and 800K VQA samples, 
coupling motion prediction with driving scene understanding.
3) NAVSIM adaptation.
The model is fine-tuned for three epochs on 103K NAVSIM trajectories to bridge the distribution gap.
4) Online reinforcement learning.
The policy is optimized using GSPO on NAVSIM data over two consecutive reinforcement learning epochs, 
with the first epoch's output serving as the reference model for the second.

\noindent\textbf{Inference.}
During inference, 
we fix the hybrid output length to 32 tokens and run 32 mask‑diffusion iterations with a monotonically decreasing mask schedule. 
Each iteration jointly infills all masked tokens and re‑masks the lowest‑confidence subset according to the selected policy (default confidence‑based; causal and reverse‑causal are optional), 
terminating early when no masks remain. 
We employ fast‑dLLM decoding for acceleration and disable step‑skipping to avoid quality regressions.

\section{Experiments}
\label{sec:experiment}
\subsection{Experimental Setups}
All experiments were conducted on $4\times 8$ Ascend 910B NPUs across four sequential training phases.
We use AdamW for all phases with a learning rate of $1\times 10^{-5}$, 
a cosine learning-rate schedule, a warm-up ratio of 0.02, 
and weight decay of 0. 
The batch size is set to 1 for memory efficiency. 
For GSPO, we use a group size of 3. 
Our MoE module contains 64 experts, each with a LoRA rank of 32 and input/output dimensions of 4096, 
with a dropout rate of 0.05. 
We adopt expert-choice routing with a routing capacity of 0.1. 
Training is performed in $\mathrm{bf16}$ precision.

\begin{table*}[!t]
\centering
\small
\setlength{\tabcolsep}{4pt}
\resizebox{0.7\textwidth}{!}{
\begin{tabular}{@{}l|cccc|ccccc|c@{}}
\hline
\textbf{Method} & \textbf{NC}$\uparrow$ & \textbf{DAC}$\uparrow$ & \textbf{DDC}$\uparrow$ & \textbf{TLC}$\uparrow$ & 
\textbf{EP}$\uparrow$ & \textbf{TTC}$\uparrow$ & \textbf{LK}$\uparrow$ & \textbf{HC}$\uparrow$ & \textbf{EC}$\uparrow$ & \textbf{EPDMS}$\uparrow$ \\
\hline
Ego Status & 93.1 & 77.9 & 92.7 & 99.6 & 86.0 & 91.5 & 89.4 & \textbf{98.3} & 85.4 & 64.0 \\
TransFuser~\cite{prakash2021multi} & 96.9 & 89.9 & 97.8 & 99.7 & 87.1 & 95.4 & 92.7 & \textbf{98.3} & 87.2 & 76.7 \\
HydraMDP++~\cite{li2024hydra} & 97.2 & 97.5 & \textbf{99.4} & 99.6 & 83.1 & 96.5 & 94.4 & 98.2 & 70.9 & 81.4 \\
DriveSuprem~\cite{yao2025drivesuprim} & 97.5 & 96.5 & \textbf{99.4} & 99.6 & \textbf{88.4} & 96.6 & 95.5 & \textbf{98.3} & 77.0 & 83.1 \\
ARTEMIS~\cite{feng2025artemis} & 98.3 & 95.1 & 98.6 & 99.8 & 81.5 & 97.4 & 96.5 & \textbf{98.3} & - & 83.1 \\
DiffusionDrive~\cite{liao2025diffusiondrive} & 98.2 & 95.9 & \textbf{99.4} & 99.8 & 87.5 & 97.3 & \textbf{96.8} & \textbf{98.3} &\textbf{ 87.7} & 84.5 \\
DriveVLA-W0\cite{li2025drivevla} & 98.5 & \textbf{99.1} & 98.0 & 99.7 & 86.4 & 98.1 & 93.2 & 97.9 & 58.9 & 86.1 \\
\hline
\textbf{Ours} & \textbf{99.0} & 98.4 & 99.3 & \textbf{99.9} & 87.0 & \textbf{98.6} & 96.2 & 98.1\ & 78.5 &\textbf{89.7}\\
\hline
\end{tabular}
}
\vspace{-2mm}
\caption{
Comparison with state-of-the-art methods on the NAVSIM-v2 with extended metrics.
}
\vspace{-2mm}
\label{tab:navsim_v2}
\end{table*}

\begin{figure*}[!t]
\centering
\begin{minipage}[t]{0.56\textwidth}
\centering
    \small
    \setlength{\tabcolsep}{4pt}
    \resizebox{\linewidth}{!}{
    \begin{tabular}{l|c|c|c|c|c|c|c|c|c|c}
    \hline
    \textbf{ID}&\textbf{DS}&\textbf{CFG}&\textbf{MOE} & \textbf{GSPO} & \textbf{NC}$\uparrow$ & \textbf{DAC}$\uparrow$ & \textbf{TTC}$\uparrow$ & \textbf{Comf.}$\uparrow$ & \textbf{EP}$\uparrow$ & \textbf{PDMS}$\uparrow$ \\
    \hline
    1 & \ding{55} & \ding{55} & \ding{55} & \ding{55} &97.0  &93.1  &91.6&99.5&76.0 &80.3\\
    2 & \checkmark & \ding{55} & \ding{55} & \ding{55} &97.0  &93.1  &91.6&99.5&76.0 &82.3\\
    3 & \checkmark & \checkmark & \ding{55} & \ding{55} &97.8&94.2&93.4&99.7&78.5&84.7  \\
    4 & \checkmark & \checkmark & \checkmark & \ding{55}&98.0&95.5&94.2&99.4&80.7&86.6  \\
    5 & \checkmark & \checkmark & \checkmark & \checkmark &\textbf{99.1}  &\textbf{98.3}  &\textbf{96.5}&\textbf{99.9}   &\textbf{84.4}&\textbf{91.0} \\
    \hline
    \end{tabular}
    }
    \vspace{-1mm}
    \captionof{table}{Ablation study on the proposed components of \model. 
    We evaluate the effect of decoding scheduler(DS), CFG, MoE and GSPO on NAVSIM-v1 evaluation.}
    \label{tab:ablation_all}
    \vspace{0.8mm}

    \small
    \setlength{\tabcolsep}{4pt}
    \resizebox{\linewidth}{!}{
    \begin{tabular}{l|c|c|c|c|c|c|c|c|c}
\hline
\textbf{ID} &\textbf{TTC}&\textbf{Comf}.&\textbf{EP}& \textbf{NC}$\uparrow$ & \textbf{DAC}$\uparrow$ & \textbf{TTC}$\uparrow$ & \textbf{Comf}.$\uparrow$ & \textbf{EP}$\uparrow$ & \textbf{PDMS}$\uparrow$\\
\hline
1&\checkmark&\ding{55}&\ding{55}&\textbf{99.1}  &98.2  &\textbf{96.7}&\textbf{99.9}  &84.0&90.7\\
2&\ding{55}&\checkmark&\ding{55} &99.0  &98.1  &96.2   &\textbf{99.9}&84.3   &90.6 \\
3&\ding{55}&\ding{55}&\checkmark&98.6  &97.9  &95.4   &99.4  &\textbf{84.9}&90.3 \\
4&\checkmark&\checkmark&\checkmark&\textbf{99.1}  &\textbf{98.3}  &96.5&\textbf{99.9}   &84.4&\textbf{91.0} \\
\hline
\end{tabular}
    }
    \vspace{-1mm}
    \captionof{table}{Ablation study on different reward setting.}
    \label{tab:gspo_ablation_pdms}
    \vspace{-6mm}
\end{minipage}
\hfill
\begin{minipage}[t]{0.43\textwidth}
    \centering
    \vspace{-1.1cm}
    \small
    \setlength{\tabcolsep}{4pt}
    \resizebox{\linewidth}{!}{
    \begin{tabular}{l|c|c|c|c|c|c}
    \hline
    \textbf{Method} & \textbf{NC}$\uparrow$ & \textbf{DAC}$\uparrow$ & \textbf{TTC}$\uparrow$ & \textbf{Comf}.$\uparrow$ & \textbf{EP}$\uparrow$ & \textbf{PDMS}$\uparrow$\\
    \hline
    Random   & 98.9  & 97.8  & \textbf{96.5}   & \textbf{99.9}   & 83.0 & 90.0 \\
    Causal        & 98.7  & 97.0  & 95.5   & 99.8   & 82.3 & 88.9 \\
    R. Causal& \textbf{99.1}  & \textbf{98.3}  & \textbf{96.5}   & \textbf{99.9}   & \textbf{84.4} & \textbf{91.0} \\
    \hline
    \end{tabular}
    }
    \vspace{1 mm}
   \captionof{table}{Ablation study of mask decoding schedules: Random, Causal, and Reverse Causal.}
   \vspace{-1mm}
   \vspace{6mm}
    \label{tab:ablation_ds}
    
    \small
    \setlength{\tabcolsep}{4pt}
    \resizebox{\linewidth}{!}{
    \begin{tabular}{l|c|c|c|c|c|c}
\hline
\textbf{CFG} & \textbf{NC}$\uparrow$ & \textbf{DAC}$\uparrow$ & \textbf{TTC}$\uparrow$ & \textbf{Comf}.$\uparrow$ & \textbf{EP}$\uparrow$ & \textbf{PDMS}$\uparrow$\\
\hline
w/o&97.0  &93.1  &91.6&99.5&76.0 &82.3\\
1&97.2  &93.6  &92.0  &99.4&76.9&83.2 \\
3&97.6  &93.6  &92.7&99.6&77.4&83.6 \\
5&97.7&94.0&93.4&\textbf{99.7}&78.0&84.4 \\
7&\textbf{97.7}&\textbf{94.2}&\textbf{93.4}&99.6&\textbf{78.4}&\textbf{84.7} \\
\hline
\end{tabular}
    }
    \vspace{-1 mm}
    \captionof{table}{Analysis of classifier free guidance scales.}
    \label{tab:cfg_ablation}
    \vspace{-6mm}
\end{minipage}
\end{figure*}

\subsection{Comparison With Existed Works on NAVSIM}
We evaluate \model\space on the NAVSIM-v1 (v1.1 codebase version for evaluation) and v2 (v2.2 codebase version) benchmarks, 
comparing against leading autoregressive and diffusion-based VLAs for autonomous driving. 
As shown in Table~\ref{tab:navsim_v1}, our method achieves a state-of-the-art PDMS of 91.0 on NAVSIM-v1, 
outperforming the strong diffusion-based baseline DiffusionDrive~\cite{liao2025diffusiondrive} by +2.9 points and surpassing autoregressive methods including ReCogDrive~\cite{li2025recogdrive} (+0.2) and DriveVLA-W0~\cite{li2025drivevla} (+0.8). 
The performance gains are consistent across key sub-metrics, particularly in collision avoidance (NC) and drivable area compliance (DAC), underscoring the safety benefits of our approach.

On the more comprehensive NAVSIM-v2 benchmark (Table~\ref{tab:navsim_v2}), 
which introduces additional metrics for traffic rule compliance and comfort, \model\space achieves an EPDMS of 89.7. 
This represents a significant +5.2 point improvement over DiffusionDrive~\cite{liao2025diffusiondrive} and a +3.6 point advantage over DriveVLA-W0~\cite{li2025drivevla}. 
The results demonstrate the robustness of our masked diffusion framework, 
particularly in complex driving scenarios requiring long-horizon reasoning and strict compliance with traffic regulations.

In Figure~\ref{fig:vis_compare}, 
we further demonstrate the qualitative comparison with two open-sourced models, 
DiffusionDrive~\cite{liao2025diffusiondrive} and TransFuser~\cite{chitta2022transfuser} in diverse driving scenarios.

\subsection{Comparison With Existed Works on nuScenes}
\label{sec:nuScenes}
We evaluate our method on the nuScenes dataset~\cite{nuScenes-dataset-2020} following the NAVSIM benchmark protocol~\cite{dauner2024navsim, cao2025pseudo}, which prioritizes collision rate as the primary metric. This emphasis is motivated by prior findings in NAVSIM showing that open-loop L2 distance exhibits minimal correlation with closed-loop performance. As presented in Table~\ref{tab:nuScenes}, our method achieves an average collision rate of \textbf{0.11\%} under the ST-P3 metrics, matching the best-performing non-VLA model (UniAD). More importantly, under the more comprehensive UniAD evaluation protocol, DFM-Drive achives the lowest average collision rate (\textbf{0.28\%}) among all VLA methods. 
\begin{table}[!t]
    \centering
    \resizebox{\linewidth}{!}{ 
    \renewcommand{\arraystretch}{1.2}
    \begin{tabular}{lcccc|cccc}
    \toprule
     \multirow{3}{*}{\textbf{Method}} & \multicolumn{4}{c}{\textbf{ST-P3 metrics}} & \multicolumn{4}{c}{\textbf{UniAD metrics}} \\
    \cmidrule(lr){2-5} \cmidrule(lr){6-9}
    & \multicolumn{4}{c}{\textbf{Collision (\%) ↓}} & \multicolumn{4}{c}{\textbf{Collision (\%) ↓}} \\
    \cmidrule(lr){2-5} \cmidrule(lr){6-9}
    & 1s & 2s & 3s & Avg.& 1s & 2s & 3s & Avg. \\
    \midrule
    ST-P3~\cite{ST-P3-2022}  & 0.23 & 0.62 & 1.27 &   0.71 & - & - & - &   -  \\ 
    
    Ego-MLP~\cite{Ego-MLP-2024} & 0.21 & 0.35 & 0.58 &   0.38 & - & - & - &   -\\

    InsightDrive~\cite{InsightDrive-2025} & 0.09 & 0.10 & 0.27 & 0.15 & 0.08 & \textbf{0.15} & 0.84 &  0.36 \\
    
    VAD-v2~\cite{VADv2-2024}  & 0.07 & 0.10 & 0.24 & 0.14 & - & - & - &   -  \\

    UniAD~\cite{UniAD-2023}  & \textbf{0.04} & \textbf{0.08} & 0.23 & 0.12 & 0.05 & 0.17 & 0.71 &   0.31 \\

    \midrule

    DriveVLM~\cite{tian2024drivevlm} & 0.10 & 0.22 & 0.45 &   0.27 & - & - & - &   -  \\

    GPT-Driver \cite{GPT-Driver-2023} & \textbf{0.04} & 0.12 & 0.36 &   0.17 & 0.07 & \textbf{0.15} & 1.10 & 0.44 \\
    
    AutoVLA~\cite{zhou2025autovla} & 0.13 & 0.18 & 0.28 & 0.20 & 0.14 & 0.25 & \textbf{0.53} &  0.31  \\

    DME-Driver~\cite{DME-Driver-2025} & - & - & - &   - & 0.05 & 0.28 & 0.55 &  0.29 \\
    
    \midrule
    
    \textbf{Ours}&\textbf{0.04}&0.09&\textbf{0.22}&\textbf{0.11}&\textbf{0.02}&0.17&0.66&\textbf{0.28}  \\
    
    \bottomrule
    \end{tabular}}
    \caption{
        End-to-end motion planning performance on the nuScenes~\cite{nuScenes-dataset-2020} dataset.
    }
    \vspace{-3mm}
    \label{tab:nuScenes}
\end{table}

\subsection{Ablation Studies}
\noindent\textbf{Architecture Designs.}
We conduct a comprehensive ablation study to evaluate the contribution of each core component in \model, with results summarized in Table~\ref{tab:ablation_all}. The baseline mask-diffusion model, trained solely on 668k nuPlan trajectory samples, achieves a PDMS of 80.2. 
Adding our proposed reverse-causal decoding scheduler yields an additional +2.0 PDMS, improving the stability of the generative process.
Introducing classifier-free guidance results in another +2.4 PDMS gain.
Incorporating the LoRA-MoE layer and jointly training on both VQA and trajectory data provides a further +1.9 PDMS, highlighting the benefit of multi-task learning for motion planning.  Finally, applying GSPO-based reinforcement learning contributes a substantial +5.3 PDMS, optimizing the policy for closed-loop driving performance and culminating in an overall best score of 91.0 PDMS.

\begin{figure}[!t]
  \centering
  \includegraphics[width=\linewidth]{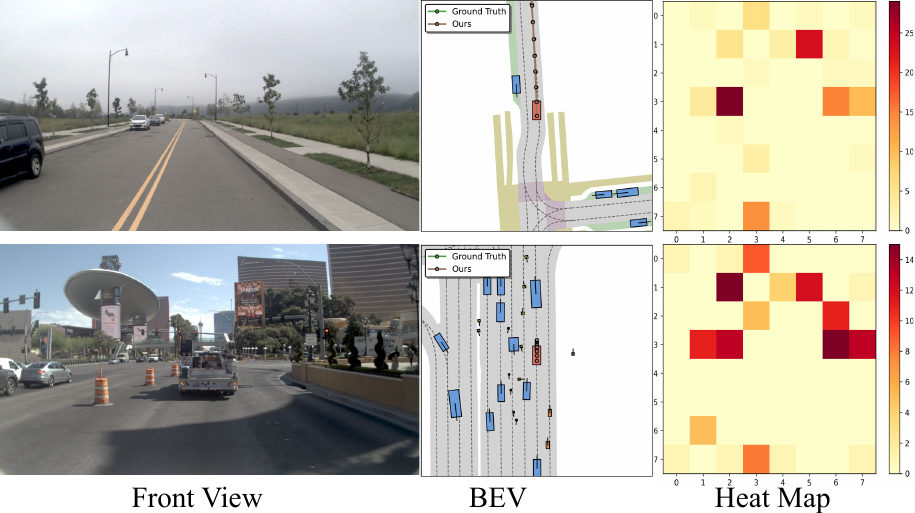} 
   \vspace{-6mm}
  \caption{Qualitative analysis of the MoE component through BEV visualizations of motion planning trajectories and corresponding expert activation heatmaps.}
   \vspace{-3.25mm}
  \label{fig:scenes_analysis_moe}
\end{figure}

\begin{figure}[!t]
  \centering
  \includegraphics[width=\linewidth]{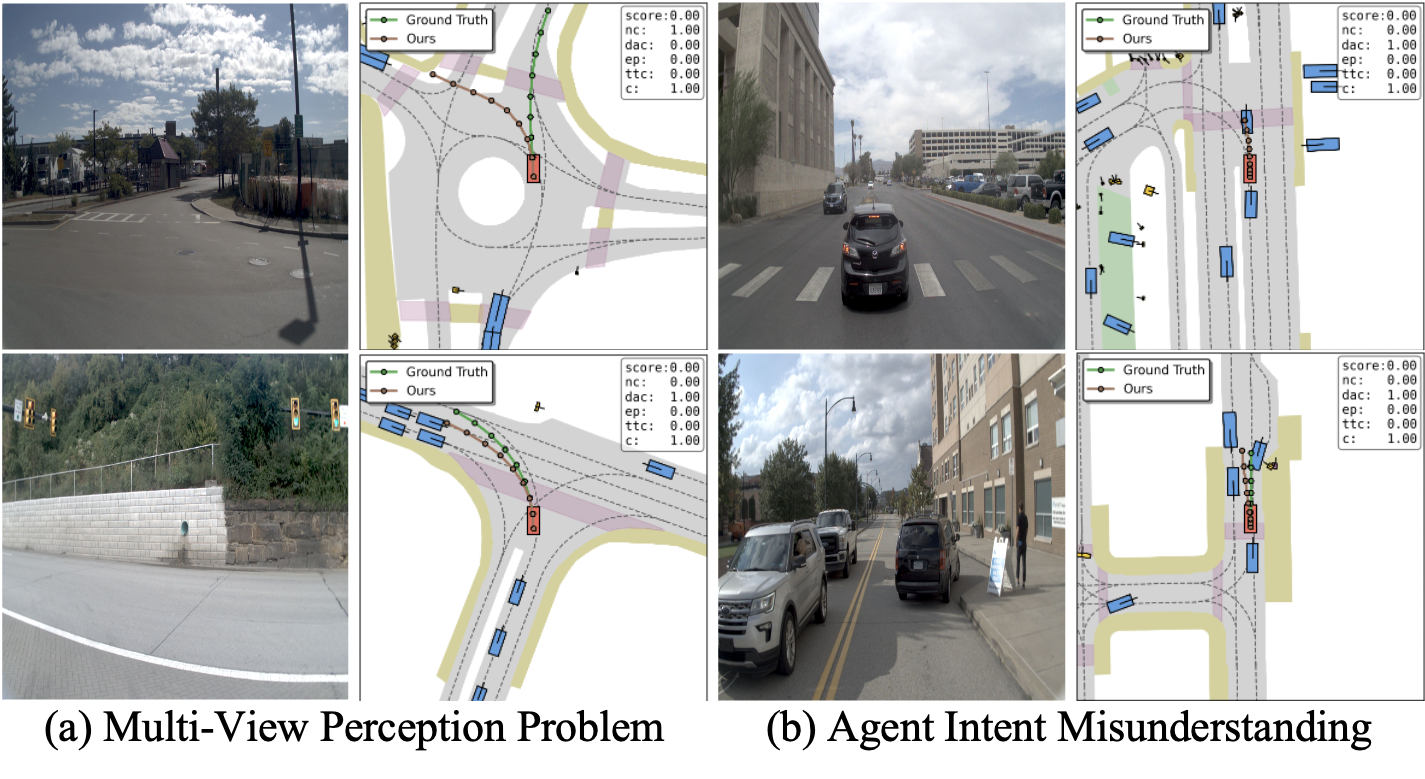} 
   \vspace{-7mm}
  \caption{Qualitative analysis of failure cases. }
  \label{fig:failure_case_vis}
  \vspace{-6mm}
\end{figure}

\noindent\textbf{Mixture-of-Experts.}
Table~\ref{tab:moe_ablation} analyzes the impact of MoE configuration on model performance. 
Without MoE, the baseline achieves a PDMS of 84.7. 
Introducing 16 experts improves performance to 85.0, and scaling to 64 experts yields a further gain of +1.6 PDMS, 
indicating that increased expert specialization enhances planning capability. 
We also evaluate the effect of LoRA rank. 
A rank of 32 yields the best performance (86.6 PDMS), 
outperforming ranks of 8 (85.5). 
Consequently, we adopt 64 experts with rank 32 as the default configuration.
Figure~\ref{fig:scenes_analysis_moe} qualitatively analyze the MoE performance on different driving scenarios.


\noindent\textbf{Reinforcement Learning GSPO.}
Tables~\ref{tab:gspo_ablation} analyze the effects of different group sizes in the proposed GSPO framework.
As shown in Table~\ref{tab:gspo_ablation}, 
introducing GSPO significantly improves model performance, 
raising PDMS from 86.6 (without GSPO) to 88.9 with a group size of 2, and further to 91.0 with a group size of 3, indicating that larger cooperative groups promote more stable and consistent policy optimization.

\begin{figure}[!t]
    \centering
    \includegraphics[width=\linewidth]{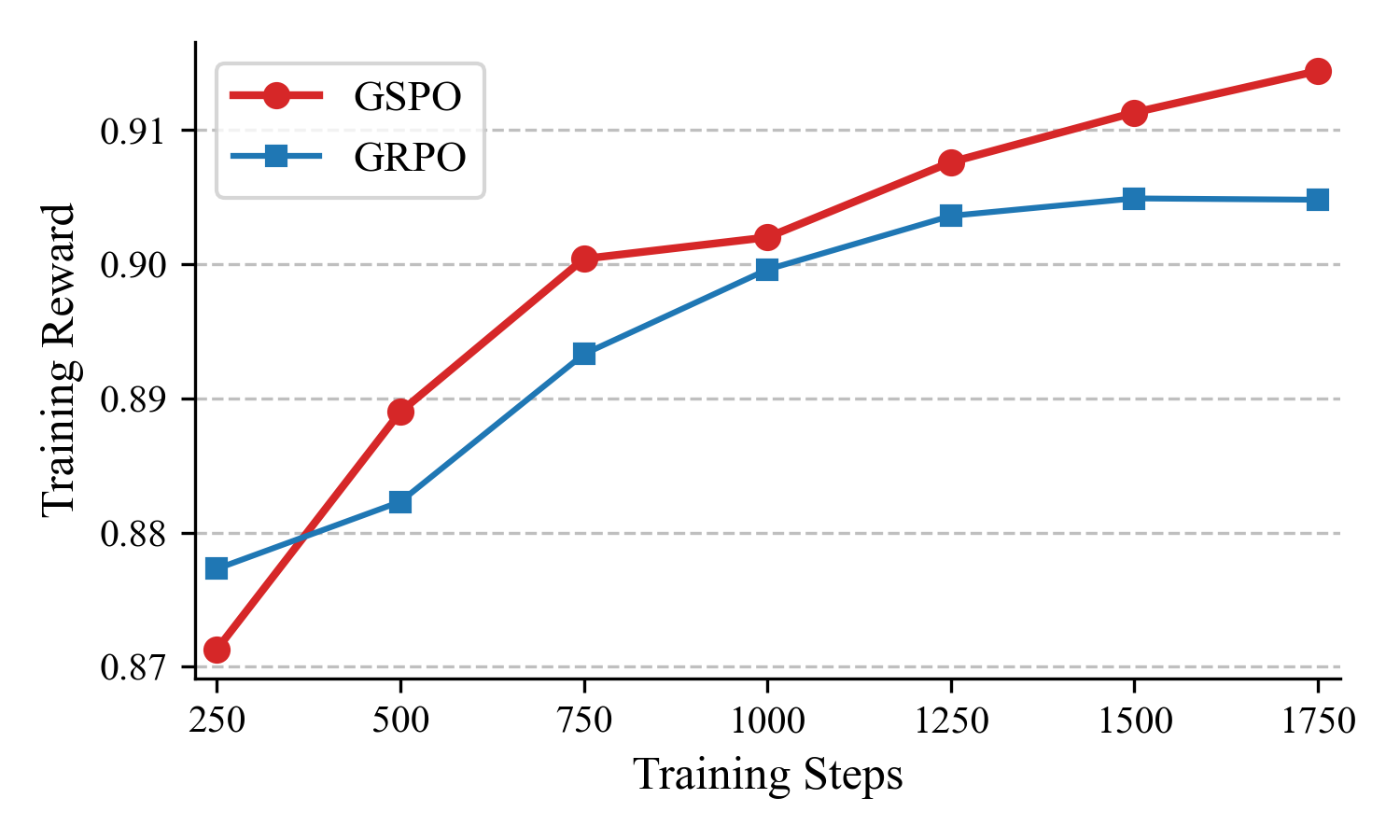}
    \vspace{-6mm}
    \caption{Training curves of reinforcement learning. GSPO possesses steady higher reward than GRPO.}
    \vspace{-3mm}
    \label{fig:gprogspo}
\end{figure}

Figure~\ref{fig:gprogspo} shows that the training reward curve with GSPO and GRPO. 
We observe that GSPO can
deliver continuous performance improvement through increasing the training compute.

\begin{figure}[!t]
  \centering
  \includegraphics[width=\linewidth]{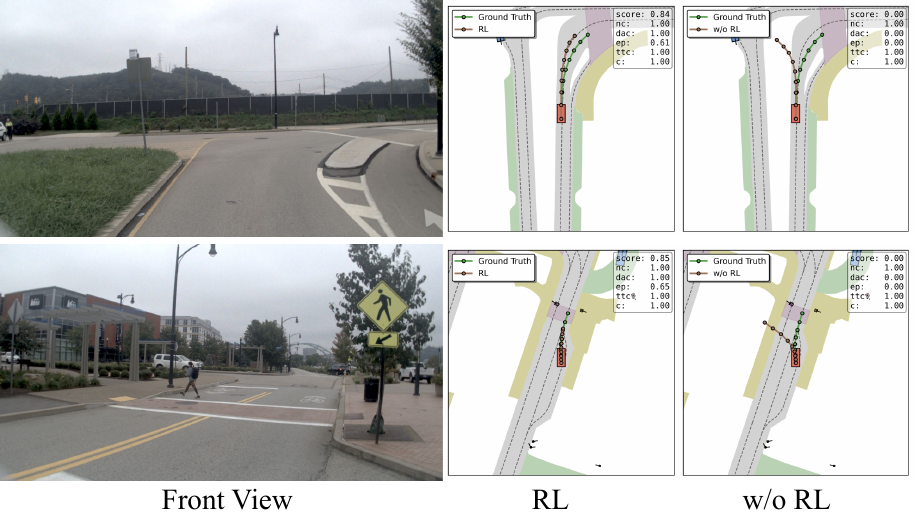} 
  \vspace{-6mm}
  \caption{Qualitative ablation of online reinforcement learning GSPO on different driving scenarios.}
   \vspace{-0.5mm}
  \label{fig:vis_compare_rl}
\end{figure}

\begin{figure}[!t]
  \centering
  \includegraphics[width=\linewidth, height=4.5cm]{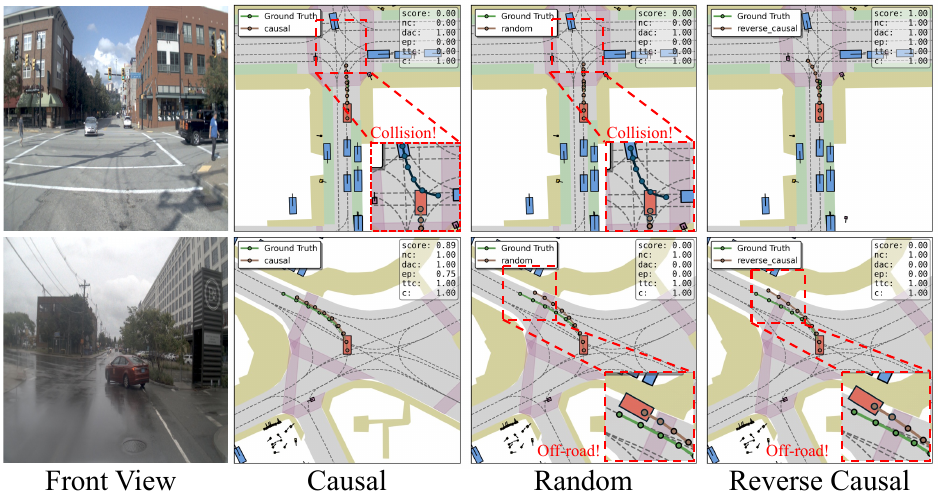} 
  \vspace{-6mm}
  \caption{Analysis of decoding orders on different scenarios.}
  \vspace{-3mm}
  \label{fig:scenes_analysis_moe_decode_schedule}
\end{figure}

Table~\ref{tab:gspo_ablation_pdms} examines reward function design. 
When only sub-scores (TTC, Comfort, EP) are preserved, 
their respective metrics improve; 
however, the overall PDMS score decreases, 
indicating over-optimization toward specific objectives. 
In contrast, using the full reward settings achieves the best balance across all dimensions, leading to the highest overall driving performance (91.0 PDMS).
As shown in Figure~\ref{fig:vis_compare_rl}, incorporating GSPO further improves trajectory generation, yielding smoother behaviors and more reliable performance in complex scenes.

\noindent\textbf{Classifier-Free Guidance.}
The impact of classifier-free guidance is evaluated in Table~\ref{tab:cfg_ablation}. 
Without CFG, the model achieves a PDMS of 82.3. 
Applying CFG consistently improves performance across guidance scales, 
with PDMS increasing to 84.7 at scales of 7. 
This improvement indicates that moderate guidance enhances planning stability and trajectory confidence.

\noindent\textbf{Masked Diffusion Decoding Scheduler.}
Table~\ref{tab:ablation_ds} evaluates the effect of decoding order on \model\space performance. 
A random decoding schedule achieves a PDMS of 90.0, 
while a causal schedule slightly reduces performance to 88.9. 
The reverse-causal schedule yields the highest PDMS of 91.0, 
with consistent gains across all metrics, 
including a +1.4 improvement in EP and +1.3 in DAC. 
As shown in Figure~\ref{fig:scenes_analysis_moe_decode_schedule}, 
causal decoding is particularly effective for turning scenarios, 
reverse-causal excels in complex interactions such as car-following and oncoming traffic, 
and random scheduling provides balanced performance across diverse scenarios.

\subsection{Limitations and Future Works}
\label{subsec:limitations}
Although \model\space achieves state-of-the-art performance on both the NAVSIM v1 and v2 benchmarks, it still has several limitations. Figure~\ref{fig:failure_case_vis} illustrates two representative failure cases. First, due to computational constraints, our model currently receives only a front-view image as input, which leads to perception failures when important obstacles lie outside this field of view. Second, the model processes only the current frame without any temporal history, making it difficult to infer other agents’ motion patterns and intent, potentially resulting in suboptimal or unsafe planning decisions.
Future work may address these issues by designing a 3D vision encoder that better aligns with textual features and by developing more efficient model architectures capable of leveraging temporal information.

\section{Conclusion}
\label{sec:conclusion}
In this work, we introduced \model, a vision-language-action framework that leverages masked discrete diffusion for trajectory generation in end-to-end autonomous driving. By unifying flexible non-causal decoding, a sparse MoE architecture, and reinforcement learning via GSPO, our approach achieves competitive performance on the NAVSIM benchmarks. The results demonstrate that masked diffusion offers a powerful and scalable alternative to conventional autoregressive and continuous diffusion models, enabling more adaptive and scenario-aware planning. Future research will explore extending this framework to larger-scale real-world datasets and incorporating richer multimodal reasoning for enhanced decision-making in complex driving environments.

{
    \small
    \bibliographystyle{ieeenat_fullname}
    \bibliography{main}
}

\clearpage
\appendix
\setcounter{page}{1}
\maketitlesupplementary

In the supplementary materials, 
we first describe the evaluation metrics for NAVSIM v1~\cite{dauner2024navsim} and v2~\cite{cao2025pseudo} in Section~\ref{sec:v1_metrics} and \ref{sec:v2_metrics}, respectively. 
Following this, 
Section~\ref{sec:code} provides the pseudo-code for our training and inference procedures, 
and Section~\ref{sec:hyper_parameter} details the hyperparameters used in the four-stage training pipeline. 
Additionally, 
we present more qualitative comparisons with state-of-the-art methods alongside an analysis of different decoding schedules in Section~\ref{sec:qualitative}. 

\section{Implementation Details}

\subsection{NAVSIM v1 Evaluation Metrics}\label{sec:v1_metrics}

NAVSIM v1~\cite{dauner2024navsim} metrics include No at-fault Collision (NC), Drivable Area Compliance (DAC), 
Time-To-Collision (TTC), Comfort (C), and Ego Progress (EP). NAVSIM uses the Predictive Driver Model 
Score (PDMS) to evaluate model performance:
\begin{equation}
\mathrm{PDMS} = NC \times DAC \times \frac{5 \times EP + 5 \times TTC + 2 \times C}{12}.
\end{equation}

\noindent\textbf{No at-fault Collision (NC)}: Penalizes collisions based on fault assignment. NC=1 indicates no at-fault collisions, NC=0.5 indicates one fault collision with static objects, and NC=0 indicates multiple fault collisions.

\noindent\textbf{Drivable Area Compliance (DAC)}: Measures adherence to drivable areas (lanes, parking areas). DAC=1 when the ego bounding box remains entirely within drivable areas, and DAC=0 when any corner exits designated areas.

\noindent\textbf{Ego Progress (EP)}: Quantifies navigation goal achievement as the ratio of actual progress to a search-based safe upper bound derived from PDM-Closed trajectories. The ratio is clipped to [0,1], with low or negative values discarded.

\noindent\textbf{Time-to-Collision (TTC)}: Encourages maintenance of safe distances from other vehicles. TTC=1 when the minimum time-to-collision exceeds 0.9 seconds, and 0 otherwise.

\noindent\textbf{Comfort (C)}: Assesses kinematic constraints including acceleration and jerk. C=1 when all predefined thresholds are satisfied, and 0 upon any violation.

\begin{algorithm}[!t]
    \caption{Training \model}
    \begin{algorithmic}[1]
      \Require \model\space $p_\theta$, Data $p_{data}$, Mask Token $M$
      \For{$step = 1$ \textbf{to} max training step}
          \State $c_t, x_0 \sim p_{data}$
          \State $r \sim \mathcal{U}(0,1)$
          
          \State Sample mask $\mathbf{m} \sim \text{Bernoulli}(r)$
          \State $x_r \leftarrow \mathbf{m} \odot M + (1-\mathbf{m}) \odot x_0$ 
          
          \State $\mathcal{L} \leftarrow - \frac{1}{r} \sum_{i=1}^{L} \mathbf{1}\{x_r^i = M\} \log p_\theta(x_0^i | x_r, c_t)$
          \State Update $\theta \leftarrow \theta - \eta \nabla_\theta \mathcal{L}$
      \EndFor
      \State \Return $p_\theta$
    \end{algorithmic}
    \label{algo:train}
\end{algorithm}
    
\begin{algorithm}[!t]
  \caption{Inference}
  \begin{algorithmic}[1]
    \Require \model\space $p_\theta$, Context $c_t=\{I_t,s_t,u_t\}$
            , Sampling Step $N$, Action Sequence Length $L$, Decoding Scheduler $R_{ds}$
    \State $x_{1} \gets M^{L}$ \hfill \# fully masked sequence
    \For{$r\gets 1$ \textbf{down to } $1/N$ \textbf{step} $1/N$}
      \State $s=1-1/N$
      \State $x_0=\arg\max_{x_0}p_\theta(x_0 \mid x_{r}, c_t)$
      \For{$i \gets 0$ \textbf{to} $L$}
            \If{$x_r^i\neq M$}
                \State $x_0^i=x_r^i$
            \Else
                \State $x_0^i=R_{ds}(x_r^i,M)$ 
            \EndIf
      \EndFor
      \State $x_r=x_0$
    \EndFor
    \State \Return $x_0$
  \end{algorithmic}
  \label{algo:inf}
\end{algorithm}

\begin{table}[!h]
\centering
\resizebox{\linewidth}{!}{
\begin{tabular}{llc}
\toprule
\textbf{Stage} & \textbf{Parameter} & \textbf{Value} \\
\midrule
\multirow{6}{*}{I} & Number of epochs & 0.2 \\
& Batch size & 1 \\
&Dataset&nuPlan~(668k)\\
& Learning rate & $1e^{-5}$ \\
& Weight decay & 0 \\
& Warmup ratio & 0.02 \\
& Learning rate schedule & Cosine \\
\midrule
\multirow{6}{*}{II} & Number of epochs & 1 \\
& Batch size & 1 \\
&Dataset&nuPlan~(668k)+VQA~(800k)\\
& Learning rate & $1e^{-5}$ \\
& Weight decay & 0 \\
& Warmup ratio & 0.02 \\
& Learning rate schedule & Cosine \\
\midrule
\multirow{6}{*}{III} & Number of epochs & 3 \\
& Batch size & 1 \\
&Dataset&NAVSIM~(103k)\\
& Learning rate & $1e^{-5}$ \\
& Weight decay & 0 \\
& Warmup ratio & 0.02 \\
& Learning rate schedule & Cosine \\
\midrule
\multirow{6}{*}{IV} & Number of epochs & 2 \\
& Batch size & 1 \\
&Dataset&NAVSIM~(103k)\\
&Group Size&3\\
& Learning rate & $1e^{-5}$ \\
& Weight decay & 0 \\
& Warmup ratio & 0.02 \\
& Learning rate schedule & Cosine \\
\bottomrule
\end{tabular}
}
\caption{Hyperparameters for \model.}
\label{tab:hp}

\end{table}

\subsection{NAVSIM v2 Evaluation Metrics}\label{sec:v2_metrics}
NAVSIM v2~\cite{cao2025pseudo} includes several components, categorized as penalties 
or weighted subscores. Key metrics are No at-fault Collision (NC), Drivable Area Compliance (DAC), 
Driving Direction Compliance (DDC), Traffic Light Compliance (TLC), Ego Progress (EP), Time to 
Collision (TTC), Lane Keeping (LK), History Comfort (HC), and Extended Comfort (EC). NAVSIM v2 uses 
the Extended Predictive Driver Model Score (EPDMS) to evaluate model performance:

\begin{equation}
\begin{aligned}
\mathrm{EPDMS} &= NC \times DAC \times DDC \times TLC \\
&\quad\times \frac{5EP + 5TTC + 2LK + 2HC + 2EC}{16}.
\end{aligned}
\end{equation}

\noindent \textbf{Driving Direction Compliance (DDC)}: 
Penalizes reverse driving behavior. 
DDC=1 for reverse distance $<2$m, DDC=0.5 for $2-6$m, 
and DDC$=0$ for $>6$m.

\noindent \textbf{Traffic Light Compliance (TLC)}: 
Measures obedience to traffic signals. 
TLC$=1$ when traffic rules are followed, 
and 0 upon violations.

\noindent \textbf{Lane Keeping (LK)}: 
Evaluates lateral positioning relative to lane centerlines, 
scored continuously from 0 to 1.

\noindent \textbf{History Comfort (HC)}: 
Assesses trajectory consistency with historical motion patterns, 
ranging from 0 to 1.

\noindent \textbf{Extended Comfort (EC)}: 
Compares planned trajectories across consecutive frames for dynamic consistency, 
scored from 0 to 1.



\subsection{Pseudo-Code of Training and Inference}\label{sec:code}
In this section, we present the training and inference algorithms. Specifically, we introduce the training and inference algorithms in Algorithm~\ref{algo:train} and Algorithm \ref{algo:inf}, respectively.

\subsection{Training Hyperparameter.}\label{sec:hyper_parameter}
Training proceeds in four stages, summarized in Table~\ref{tab:hp}. 
Stage~I initializes the MoE by freezing the backbone and training only the LoRA experts for 0.2~epochs on 668K nuPlan trajectories. 
Stage~II performs full-parameter multi-task learning for 1~epoch on 668K nuPlan trajectories combined with corresponding 800K VQA samples. 
Stage~III adapts the model to the NAVSIM domain through 3~epochs of supervised fine-tuning on 103K trajectories. 
Stage~IV conducts 2~epochs of NAVSIM training with a group size of 3 to support multi-scenario conditioning. 
Across all stages, we use a batch size of~1, a learning rate of $1\times10^{-5}$, a warmup ratio of 0.02, zero weight decay, and a cosine learning-rate schedule.

\section{Additional Experiment Results}

\subsection{Additional Qualitative Results}\label{sec:qualitative}
\begin{figure*}[!t]
    \centering
    \includegraphics[width=\textwidth]{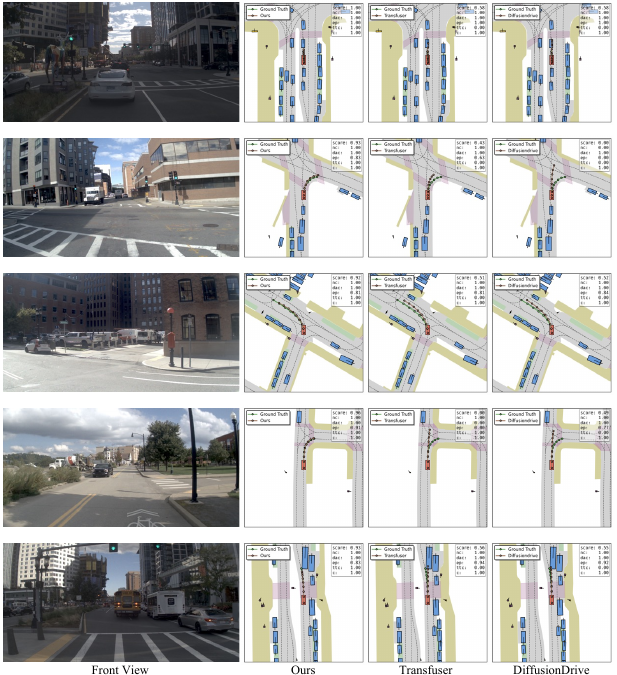}
    \caption{Qualitative results compare with existed methods.}
    \label{fig:qual_res1}
\end{figure*}

\begin{figure*}[!t]
    \centering
    \includegraphics[width=\textwidth]{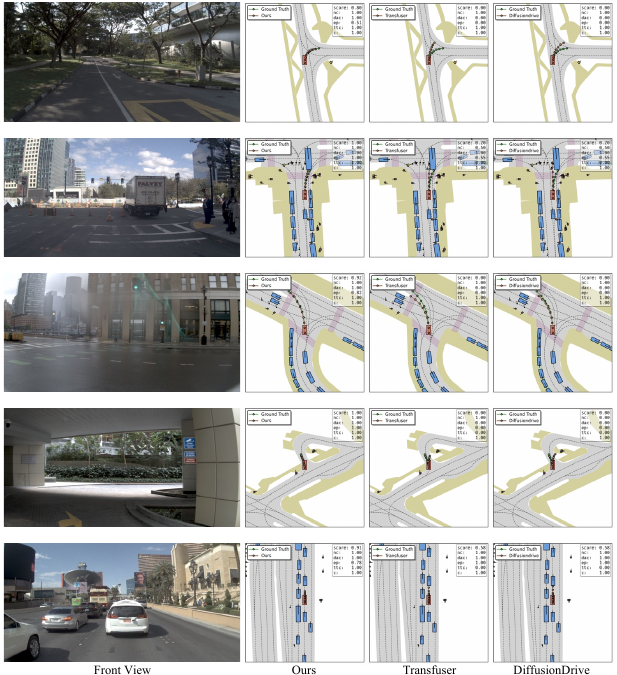}
    \caption{Qualitative results compare with existed methods.}
    \label{fig:qual_res2}
\end{figure*}

\begin{figure*}[!t]
    \centering
    \includegraphics[width=\textwidth]{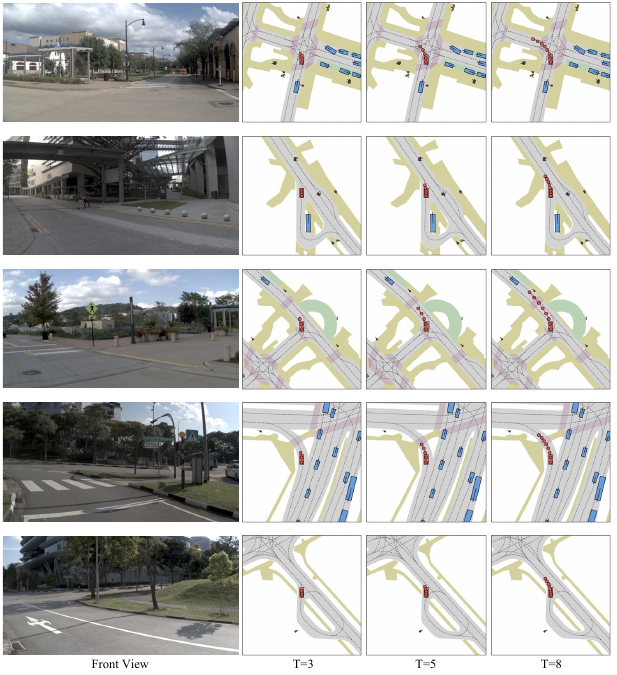}
    \caption{Qualitative results of \textbf{causal} schedule.}
    \label{fig:qual_caus}
\end{figure*}

\begin{figure*}[!t]
    \centering
    \includegraphics[width=\textwidth]{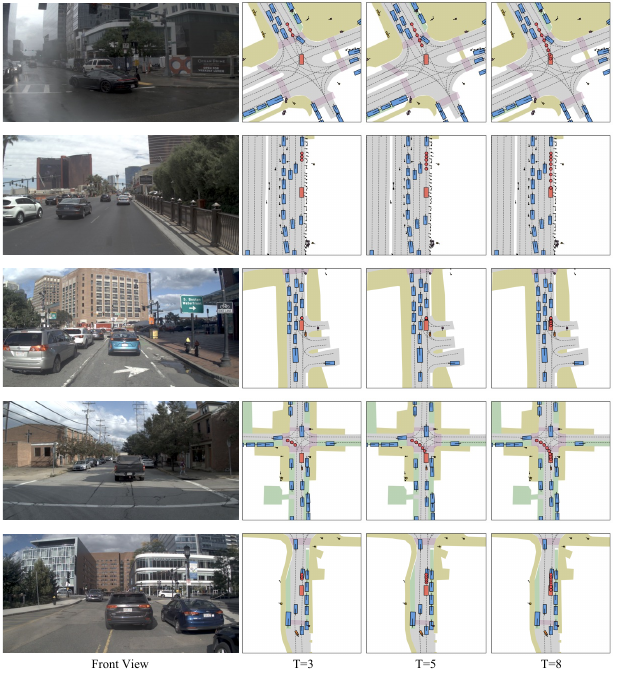}
    \caption{Qualitative results of \textbf{reverse causal} schedule.}
    \label{fig:qual_r_caus}
\end{figure*}

We present extensive qualitative visualizations of \model\space on NAVSIM comprising with Transfuser and DiffusionDrive to demonstrate the effectiveness of our proposed
method~(see Figure~\ref{fig:qual_res1} and \ref{fig:qual_res2}). We further showcase more examples of decoding orders on different scenarios~(see Figure~\ref{fig:qual_caus} and \ref{fig:qual_r_caus}).

\end{document}